\documentclass{article} 

\usepackage{natbib}
\usepackage{times}
\usepackage{makecell}

\usepackage{amsmath,amsfonts,bm}

\def\eqref#1{equation~\ref{#1}}

\def\1{\bm{1}}

\DeclareMathAlphabet{\mathsfit}{\encodingdefault}{\sfdefault}{m}{sl}
\SetMathAlphabet{\mathsfit}{bold}{\encodingdefault}{\sfdefault}{bx}{n}

\usepackage{hyperref}
\usepackage{url}
\usepackage{algorithm}
\usepackage[indLines=true]{algpseudocodex}
\usepackage{dsfont}
\usepackage{multirow}
\usepackage{amssymb}
\usepackage{amsmath}

\algrenewcommand\alglinenumber[1]{\scriptsize #1\hspace{0.6em}}

\title{Event-Driven Parameter Dynamics for Spatiotemporal Parameter-Update Application in Deep Learning}

\author{ Ruixue Li \\
School of Electronics and Information Engineering \\
Hangzhou Dianzi University \\
Hangzhou, China \\
\texttt{rxli2020@hdu.edu.cn} \\
\and
Chaofei Hong \\
Zhejiang Lab \\
Hangzhou, China \\
\texttt{hongchf@zhejianglab.org} \\
}

\begin{document}

\maketitle
\begin{abstract}
Standard neural network training typically applies optimizer-generated update signals to model parameters at every iteration. This work introduces a spatiotemporal view of parameter updating and proposes Event-Driven Parameter Dynamics (EPD), which separates candidate-update generation from their application through an intermediate state with accumulation, event-triggered release, and reset. In a canonical coordinate-wise instantiation, accumulation produces temporal integration, release imposes a spatial budget, and reset separates successive inter-release intervals, while conventional dense updating is recovered as a boundary case. Across four model--optimizer configurations on CIFAR-100, EPD maintains final performance close to conventional training even when only a small fraction of coordinates are released at each step, including the 1\% regime. Matched-control experiments further show that temporal accumulation, state-dependent selection, and post-release reset each contribute to the observed behavior. Together, these results identify parameter-update realization as a controllable dimension of optimization and provide a concrete basis for studying its temporal and spatial organization separately from candidate-update generation.
\end{abstract}

\section{Introduction}
\label{sec:intro}

Modern gradient-based optimizers, such as stochastic gradient descent, momentum methods, and adaptive optimizers \citep{robbins1951stochastic,polyak1964some,kingma2015adam,loshchilov2019decoupled}, primarily differ in how they numerically construct parameter updates, but typically share a dense, step-wise parameter-updating paradigm in neural network training. Under this paradigm, primary attention is directed toward the numerical properties of update values, while when and where these updates are committed to the parameter vector are largely inherited from the dense execution form rather than treated as independent design choices. In this work, we refer to this arrangement as the spatiotemporal organization of parameter updating. We hypothesize that, beyond the numerical construction of update values, the spatiotemporal organization of parameter updates can serve as an explicit degree of freedom in neural network optimization and may influence the resulting optimization behavior.

This perspective also allows several existing approaches to be reinterpreted as modifying the temporal or coordinate-wise extent of parameter updates, although they were not originally introduced from this viewpoint. A direct line of work comes from coordinate and block-coordinate optimization, where updates are applied to selected variables or parameter blocks rather than to the full parameter vector at every iteration \citep{wright2015coordinate}. Related formulations have also been studied in deep neural network training, including block-coordinate training schemes and convergence analyses for nonconvex deep models \citep{zhang2017convergent,zeng2019global,nakamura2021block}. Within deep learning, selective gradient and sparse backpropagation methods further restrict the components that participate in parameter modification at each step. Rather than applying the full back-propagated gradient to all associated parameters, these methods retain, compute, or exploit only selected gradient components, weights, or biases, thereby changing the coordinate range over which updates are realized \citep{sun2017meprop,nikdan2023sparseprop,rub2023tinyprop}. Their original motivations are typically acceleration, regularization, or computational reduction, but from the present perspective they also instantiate non-standard coordinate patterns of parameter updating.

The temporal organization of parameter updating is also altered by methods such as progressive freezing, which gradually removes layers from further optimization \citep{brock2017freezeout}, and resource-constrained on-device training, which restricts trainable layers, channels, or sub-tensors according to memory and computation budgets \citep{lin2022ondevice,kwon2024tinytrain}. Gradient sparsification with residual or error-feedback states introduces another form of temporal organization, where components omitted from an immediate sparse update are retained and can re-enter subsequent updates \citep{lin2018deep,stich2018sparsified,karimireddy2019error}. Event-triggered or lazy gradient methods likewise adapt when gradient information is recomputed, communicated, or reused \citep{chen2018lag}. Together, these approaches demonstrate that parameter updating can exhibit diverse temporal and coordinate-wise organizations beyond dense step-wise application, although such patterns are typically induced by method-specific objectives such as communication reduction, computational efficiency, or resource constraints.

These observations motivate a broader question. Existing methods instantiate particular update organizations through coordinate partitions, freezing schedules, residual compensation, or event rules, but the resulting realization patterns generally remain components of their respective algorithmic or system designs. This suggests an opportunity to abstract parameter realization itself as a unified dynamical design object, separating how candidate updates are generated from how they are accumulated, selected, and committed to model parameters. We address this question through \emph{Event-Driven Parameter Dynamics} (EPD), a state-dependent framework that organizes parameter realization through accumulation, event-triggered release, and reset.

We instantiate EPD in a canonical coordinate-wise form and analyze these three mechanisms at the mechanism level, with conventional dense updating recovered as a boundary case. Across four model--optimizer configurations on CIFAR-100, EPD maintains final performance close to conventional training even under highly selective realization, including the $1\%$ release regime. Release-rate analysis and matched-control experiments further examine the controllability of parameter realization and the functional roles of the three mechanisms.

Our contributions are summarized as follows.
\begin{itemize}
\item A spatiotemporal view of parameter updating is introduced to study how updates are arranged across optimization steps and parameter coordinates.
\item EPD is proposed as a state-dependent parameter-updating framework that
decouples candidate-update generation from parameter-update application through accumulation, event-triggered release, and reset.
\item A controlled empirical study of the canonical EPD instantiation is provided, showing that it maintains test performance close to conventional training across multiple model--optimizer configurations, while matched-control
experiments identify the contributions of temporal accumulation, state-dependent release, and post-release reset.
\end{itemize}

\section{EPD: Event-Driven Parameter Dynamics}
\label{sec}

This section presents the formal update dynamics of the formulation of Event-Driven Parameter Dynamics (EPD). The central idea is to separate the generation of optimizer updates from their realization as parameter changes. Instead of directly applying optimizer-produced increments to model parameters, EPD introduces an intermediate dynamical state that accumulates update information, triggers selective release events, and evolves through a reset mechanism.

\subsection{General Formulation}
\label{sec:genral-formulation}

Consider the training of a neural network $f_\theta$ with parameters $\theta \in \mathbb{R}^d$ at optimization step $t$. Given the mini-batch loss $\ell(\theta_t)$, the gradient is
\begin{equation}
g_t = \nabla_\theta \ell(\theta_t).
\label{eq:minibatch-gradient}
\end{equation}
A base optimizer can be represented as an operator
\begin{equation}
u_t, h_{t+1} = \mathcal{B}(g_t, h_t, \theta_t),
\label{eq:base-optimizer}
\end{equation}
where $h_t$ is the optimizer state and $u_t \in \mathbb{R}^d$ is the candidate parameter increment generated by the optimizer. The operator $\mathcal{B}$ abstracts standard optimizers used in neural network training, including stochastic gradient-based methods, momentum-based methods, and adaptive optimization methods.

In conventional training, this increment is directly realized as
\begin{equation}
\theta_{t+1} = \theta_t + u_t.
\label{eq:conventional-update}
\end{equation}

By contrast, EPD introduces an additional realization process between the candidate increment and the actual parameter update. Instead of directly applying $u_t$, EPD accumulates update information into an intermediate dynamical state, selectively realizes part of this state through event-driven release, and updates the remaining state through a reset mechanism.

Let 
$\mathcal{P}=\{\mathcal{I}_1,\ldots,\mathcal{I}_K\}$
denote a predefined partition of parameter coordinates into update units, where $\mathcal{I}_k$ represents the $k$-th update unit. They can correspond to different granularities, such as individual parameter coordinates, parameter blocks, tensors, channels, layers, or other predefined parameter groups.

For an update unit $\mathcal{I}_k$, $u_t^{(k)}$ denotes the corresponding component of the optimizer-generated candidate increment, and $s_t^{(k)}$ denotes its associated internal EPD state. Before release, the candidate increment is incorporated into this state through a general accumulation operator as

\begin{equation}
\bar{s}_{t+1}^{(k)}
=
\mathcal{F}_{\mathrm{acc}}
(s_t^{(k)},u_t^{(k)}),
\qquad
k=1,\ldots,K .
\label{eq:general-accumulation}
\end{equation}

where $\mathcal{F}_{\mathrm{acc}}$ specifies how candidate increments are incorporated into the internal state across optimization steps.

After accumulation, EPD determines which accumulated states are realized as parameter updates. Here, let $\tau_{t+1}^{(k)}$ denotes the release threshold for each unit $\mathcal{I}_k$. A release score function $\Phi$ evaluates the accumulated state, and the release mask is defined as

\begin{equation}
m_{t+1}^{(k)}
=
\mathds{1}
\left(
\Phi(\bar{s}_{t+1}^{(k)})
\geq
\tau_{t+1}^{(k)}
\right),
\qquad
k=1,\ldots,K .
\label{eq:release-mask}
\end{equation}

For the parameter subvector $\theta_t^{(k)}$ associated with update unit $\mathcal{I}_k$, the realized increment and parameter update are
\begin{align}
r_{t+1}^{(k)}
&=
m_{t+1}^{(k)}
\odot
\bar{s}_{t+1}^{(k)},
\label{eq:general-release}\\
\theta_{t+1}^{(k)}
&=
\theta_t^{(k)}
+
r_{t+1}^{(k)}.
\label{eq:epd-parameter-update}
\end{align}

And then, the internal state evolves according to the reset rule

\begin{equation}
s_{t+1}^{(k)}
=
\mathcal{R}
\left(
\bar{s}_{t+1}^{(k)},
m_{t+1}^{(k)}
\right),
\qquad
k=1,\ldots,K .
\label{eq:general-reset}
\end{equation}

where $\mathcal{R}$ specifies how the state of $\mathcal I_k$ is carried forward after release. The accumulation operator, release rule, and reset mechanism jointly define an EPD instance. This formulation decouples the generation of candidate increments from their realization as parameter updates. The base optimizer determines $u_t$, while EPD governs how these increments are accumulated, selectively released, and propagated through the internal state before being realized as parameter changes.

\subsection{Canonical Coordinate-wise Instantiation}
\label{sec:canonical-instantiation}

In this work, we consider a canonical coordinate-wise instantiation of EPD, referred to as EPD for brevity unless otherwise specified. Each scalar parameter coordinate is treated as an individual update unit, $\mathcal{I}_k=\{k\}$,($k=1,\ldots,d$). Accordingly, the EPD state, release decision, and realized increment are defined at the level of individual parameter coordinates.

At optimization step $t$, the base optimizer produces the candidate parameter increment $u_t$ according to Eq.~\ref{eq:base-optimizer}. For a coordinate-wise update unit, the corresponding candidate increment is $u_t^{(k)}=u_{t,k}$. The accumulation operator in Eq.~\ref{eq:general-accumulation} is instantiated as a decayed additive update,
\begin{equation}
\bar{s}_{t+1}^{(k)}
=
\beta s_t^{(k)}
+
u_t^{(k)},
\qquad
\beta\in[0,1],
\qquad
k=1,\ldots,d.
\label{eq:canonical-accumulation}
\end{equation}
where the decay factor $\beta$ controls the retention of the previous state across optimization steps. When $\beta=0$, the previous state does not contribute to the current accumulation; for $0<\beta<1$, its contribution is retained with decay; and when $\beta=1$, it is retained without decay.

For release, the score function in Eq.~\ref{eq:release-mask} is instantiated as the magnitude of the accumulated coordinate state,
\begin{equation}
\Phi
\left(
\bar{s}_{t+1}^{(k)}
\right)
=
\left|
\bar{s}_{t+1}^{(k)}
\right|,
\qquad
k=1,\ldots,d.
\label{eq:canonical-release-score}
\end{equation}
Thus, the release decision depends on the magnitude of the accumulated state irrespective of its sign.

The release budget is specified by a target release density $p\in(0,1]$. 
Let $\mathcal{T}_j$ denote the set of parameter coordinates belonging to the $j$-th learnable tensor, with$d_j=|\mathcal{T}_j|$ coordinates. The per-tensor release budget is
\begin{equation}
q_j
\approx
p d_j,
\qquad
1\leq q_j\leq d_j,
\label{eq:tensor-release-budget}
\end{equation}
where the approximation accounts for integer rounding.

For each tensor, the $q_j$ coordinates with the largest accumulated-state magnitudes are selected for release. 
Specifically, define
\begin{equation}
\mathcal{S}_{t+1}^{(j)}
=
\operatorname{Top}_{q_j}
\left(
\left\{
\left|
\bar{s}_{t+1}^{(k)}
\right|
\right\}_{k\in\mathcal{T}_j}
\right),
\label{eq:canonical-topq-set}
\end{equation}
where $\operatorname{Top}_{q_j}(\cdot)$ returns the indices of the $q_j$ largest entries within tensor $j$. 
The coordinate-wise release mask is then
\begin{equation}
m_{t+1}^{(k)}
=
\mathds{1}
\left(
k\in\mathcal{S}_{t+1}^{(j)}
\right),
\qquad
k\in\mathcal{T}_j.
\label{eq:canonical-release-mask}
\end{equation}

This Top-$q_j$ selection is equivalent to the threshold-based formulation in Eq.~\ref{eq:release-mask}, with the tensor-specific threshold determined by the boundary of the selected accumulated-state magnitudes.

The realized increment at coordinate $k$ is then
\begin{equation}
r_{t+1}^{(k)}
=
m_{t+1}^{(k)}
\bar{s}_{t+1}^{(k)}.
\label{eq:canonical-release}
\end{equation}
Since the coordinate-wise update units partition the parameter coordinates, these realized increments jointly define the full increment $r_{t+1}$, which is applied according to Eq.~\ref{eq:epd-parameter-update}.

After release, the reset rule in Eq.~\ref{eq:general-reset} is instantiated as a hard reset,
\begin{equation}
s_{t+1}^{(k)}
=
\left(
1-m_{t+1}^{(k)}
\right)
\bar{s}_{t+1}^{(k)},
\qquad
k=1,\ldots,d.
\label{eq:canonical-reset}
\end{equation}
Hence, the state of a released coordinate is cleared after realization, whereas the accumulated state of an unreleased coordinate is carried forward to the next optimization step. Finally, the realized increment is applied to the corresponding parameter coordinate according to Eq. ~\ref{eq:general-release} and Eq. ~\ref{eq:epd-parameter-update}.

\begin{figure}
\begin{center}
\includegraphics[width=\linewidth]{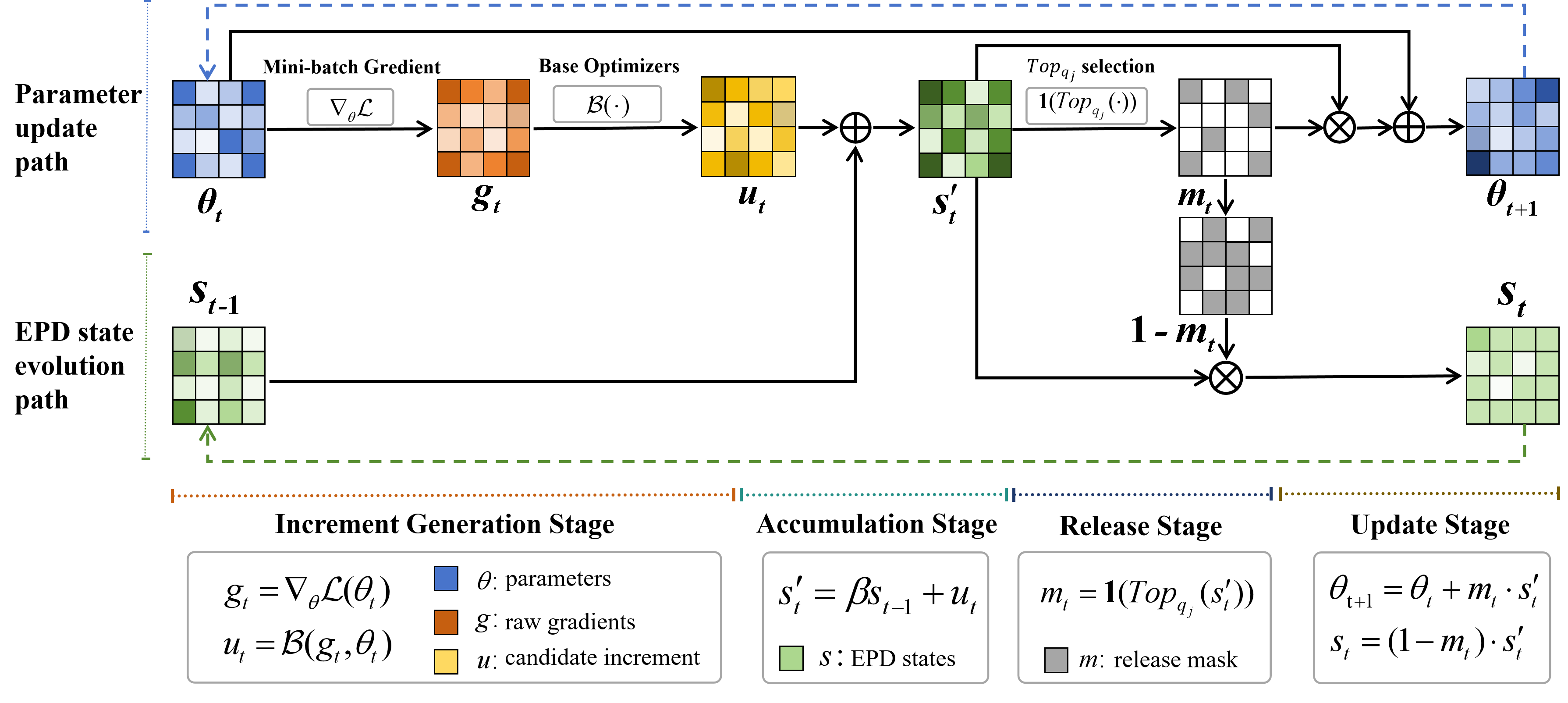}
\caption{
Illustration of the canonical coordinate-wise EPD instantiation.
The base optimizer generates the candidate increment $u_t$.
EPD accumulates the candidate increments into coordinate-wise states, selects a fixed per-tensor fraction of coordinates according to their accumulated-state magnitudes, and applies the released increments to the model parameters.
Released states are then reset, while unreleased states are carried forward.
The dashed line denotes inter-step recurrence.
}
\label{fig:canonical-epd}
\end{center}
\end{figure}

Figure~\ref{fig:canonical-epd} summarizes the coordinate-wise EPD realization used in this work. The corresponding implementation is provided in Algorithm~\ref{alg:epd-instantiation} in Appendix~\ref{app:algorithm}.

\subsection{Properties of the EPD Mechanisms}
\label{sec:epd-properties}

The accumulation, release, and reset mechanisms determine the temporal, spatial, and event-wise structure of parameter realization in EPD. General properties follow from the formulation in section ~\ref{sec:genral-formulation}, while the canonical instantiation yields the stronger results below. Detailed derivations are provided in Appendix~\ref{app:epd-properties}.

\noindent\textbf{Proposition 1.}
When the accumulation dynamics retain a nontrivial dependence on the preceding state, candidate increments from earlier steps can influence the current accumulated state. For canonical EPD, within an inter-release interval beginning from $s_{\rho}^{(k)}=0$,
\begin{equation}
\bar{s}_{t+1}^{(k)}
=
\sum_{\ell=\rho}^{t}
\beta^{\,t-\ell}u_{\ell}^{(k)} .
\label{eq:prop1-canonical}
\end{equation}
Thus, $\beta=0$ removes cross-step accumulation, $0<\beta<1$ yields exponentially decayed temporal integration, and $\beta=1$ retains preceding contributions without decay. 

\noindent\textbf{Corollary 1.}
Suppose that, within an inter-release interval,
$u_{\ell}^{(k)}=\mu^{(k)}+\varepsilon_{\ell}^{(k)}$,
where $\mu^{(k)}$ is locally stable and
$\varepsilon_{\ell}^{(k)}$ is zero-mean, mutually uncorrelated and has variance $\sigma_k^2$.
After normalization by the accumulated weights, the canonical accumulated state remains unbiased with respect to $\mu^{(k)}$ and has variance no greater than that of a single-step candidate increment.

\noindent\textbf{Proposition 2.}
In general EPD, the release mask restricts the spatial support of realized parameter updates. In canonical EPD, the per-tensor budget $q_j\approx p d_j$ further fixes how many coordinates can be realized, while their locations are determined by the accumulated-state magnitudes. Specifically,
\begin{equation}
\mathcal{S}_{t+1}^{(j)}
\in
\arg\max_{\substack{
\mathcal{S}\subseteq\mathcal{T}_j\\
|\mathcal{S}|=q_j}}
\sum_{k\in\mathcal{S}}
\left|
\bar{s}_{t+1}^{(k)}
\right|^2 .
\label{eq:prop2-topq}
\end{equation}
Hence, $p$ controls the amount of spatial realization, whereas the accumulated state determines where realization occurs. The resulting update is a best $q_j$-sparse approximation of the accumulated state within each tensor in the $\ell_2$ sense.

\noindent\textbf{Corollary 2.}
For canonical EPD with hard reset and $s_0=0$, setting $p=1$, corresponding to $100\%$ release, releases all coordinates at every step. The resulting realization reduces to the conventional dense update of the base optimizer.

\noindent\textbf{Proposition 3.}
In general EPD, the reset rule determines how post-release state information propagates into subsequent dynamics. Under canonical hard reset, the next accumulated state satisfies
\begin{equation}
\bar{s}_{t+1}^{(k)}
=
\beta
\left(1-m_t^{(k)}\right)
\bar{s}_{t}^{(k)}
+
u_t^{(k)} .
\label{eq:prop3-hard-reset}
\end{equation}
When $m_t^{(k)}=1$, the previously accumulated state is removed and $\bar{s}_{t+1}^{(k)}=u_t^{(k)}$; when $m_t^{(k)}=0$, the accumulated state continues to propagate with factor $\beta$. Hard reset therefore separates the temporal history of each coordinate into successive inter-release accumulation intervals.

\section{Experiments and Analysis}
\label{sc:experiments}

The empirical evaluation is scoped to the canonical EPD instantiation described in Section~\ref{sec:canonical-instantiation}. We evaluate EPD on CIFAR-100 using ResNet-18 and DenseNet-121 with two base optimizers, AdamW and SGD with momentum (SGD-M), resulting in four model--optimizer settings. We study its empirical effectiveness, the contributions of its main mechanisms, and the resulting event-driven parameter dynamics. Quantitative results are aggregated over 10 random seeds unless otherwise specified, with detailed training and implementation settings provided in Appendix~\ref{app:experimental-setting}.

\subsection{Main Results}
\label{sec:main-results}

Table~\ref{tab:main-results} reports the final test performance of the canonical EPD instantiation across four model--optimizer settings. Replacing conventional immediate parameter realization with the state-mediated realization dynamics of EPD results in limited changes in final accuracy across the evaluated configurations. The mean top-1 accuracy differences relative to training without EPD are $-0.92$, $-0.73$, $-0.34$, and $-1.55$ percentage points for DenseNet-121 with AdamW, DenseNet-121 with SGD-M, ResNet-18 with AdamW, and ResNet-18 with SGD-M, respectively. Across the four settings, the run-to-run variability remains of a similar magnitude to that observed in conventional training. These results indicate that the canonical EPD realization can maintain effective predictive performance even when parameter increments are no longer realized immediately at every optimization step. Detailed per-seed results are provided in Table~\ref{tab:per-seed-results} of Appendix~\ref{app:experimental-details}.

Figure~\ref{fig:main-curves}(a--d) further examines the corresponding optimization trajectories. In all four settings, EPD produces smooth and non-divergent training-loss curves and reaches a low-loss regime comparable to conventional training. The detailed trajectories vary across optimizer settings. With AdamW, the EPD and conventional training curves largely overlap, whereas with SGD-M, EPD reaches lower training loss earlier in the training process. Overall, EPD maintains effective and well-behaved optimization across both model architectures and base optimizers, providing empirical support for the viability of state-mediated, event-driven parameter realization.

\begin{table}[h]
\caption{Main results on CIFAR-100. Top-1 test accuracy is reported as mean 
\(\pm\) standard deviation over 10 random seeds. SGD-M means SGD with momentum. Here, EPD denotes the canonical coordinate-wise instantiation described in Section~\ref{sec:canonical-instantiation}. }
\label{tab:main-results}
\begin{center}
\setlength{\tabcolsep}{5pt}
\renewcommand{\arraystretch}{1.12}
\begin{tabular}{lllcc}
\hline
\multicolumn{1}{c}{\bf Model} &
\multicolumn{1}{c}{\bf Optimizer} &
\multicolumn{1}{c}{\bf Setting} &
\multicolumn{1}{c}{\bf Top-1 Acc.(\%)} &
\multicolumn{1}{c}{\bf $\Delta$ (\%)}
\\ \hline \\[-1.5ex]
ResNet-18
& AdamW 
& without EPD 
& $74.00 \pm 0.28$
& -- \\
&
& with EPD
& $73.65 \pm 0.19$
& $-0.34 \pm 0.40$ \\

\\[-0.5ex]

ResNet-18 
& SGD-M
& without EPD 
& $78.76 \pm 0.31$ 
& -- \\
&
& with EPD 
& $77.22 \pm 0.24$
& $-1.55 \pm 0.31$ \\
\\[-0.5ex]

DenseNet-121
& AdamW 
& without EPD 
& $76.91 \pm 0.22$ 
& --\\
&
& with EPD 
& $75.98 \pm 0.27$
& $-0.92 \pm 0.42$\\
\\[-0.5ex]

DenseNet-121 
& SGD-M
& without EPD 
& $80.58 \pm 0.28$
& --\\
&
& with EPD
& $79.85 \pm 0.28$
& $-0.73 \pm 0.46$\\
\hline

\end{tabular}
\end{center}
\end{table}

The target release rate $p$ directly controls the intensity of parameter realization in the canonical EPD instantiation by determining the adaptive release threshold at each step. Varying $p$ therefore provides a direct way to examine how different realization regimes affect optimization and final performance. We study this behavior using ResNet-18 with AdamW as a representative configuration, while keeping the remaining EPD settings fixed. Figure~\ref{fig:main-curves}(e) reports the resulting test performance across different values of $p$.

At the most selective release setting, $p=0.5\%$, the final accuracy decreases
more noticeably. Increasing $p$ to $1\%$ largely recovers the performance.
Beyond this point, further increases in $p$ produce comparatively small
changes, with the final accuracy remaining within a narrow range from
$p=1\%$ to $100\%$ and showing no clear monotonic dependence on the release
rate. In particular, the $1\%$ setting achieves performance close to much
denser realization regimes, indicating that effective optimization can be
maintained even when parameter realization is highly selective. Overall,
these results suggest that the canonical EPD instantiation admits a broad
effective operating range rather than relying on a narrowly tuned release
rate. Detailed per-seed results are provided by Table ~\ref{tab:release-rate-per-seed} in Appendix~\ref{app:experimental-details}.

\begin{figure}[t]
\begin{center}
\begin{minipage}[t]{1.0\textwidth}
\vspace{0.6em}
\begin{center}
\includegraphics[width=\textwidth]{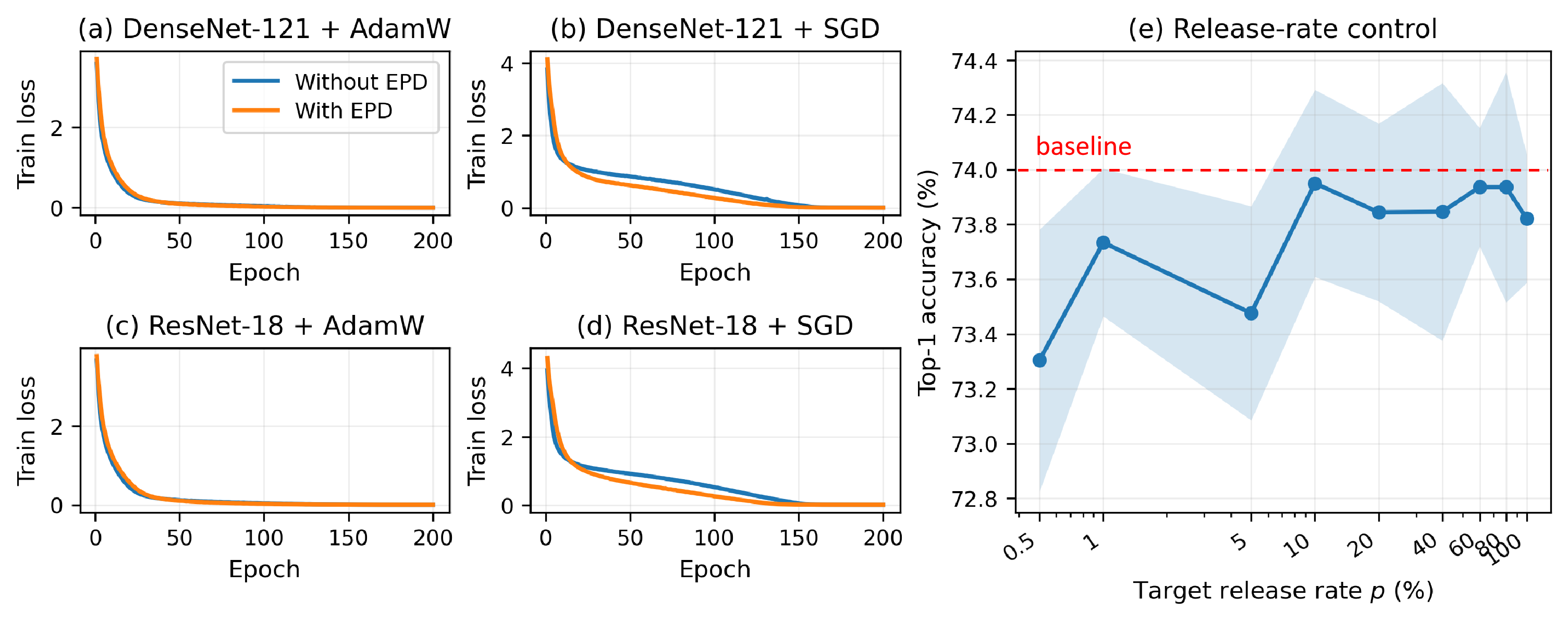}\\[-1mm]

\end{center}
\end{minipage}
\caption{
Main empirical results of the canonical EPD instantiation on CIFAR-100.
(a--d) Training-loss trajectories for the four model--optimizer settings.
Solid curves and shaded regions denote the mean and one standard deviation over 10 random seeds, respectively.
(e) Final top-1 test accuracy across different target release rates $p$ for ResNet-18 with AdamW, with the remaining EPD settings fixed.
} 
\label{fig:main-curves}
\end{center}
\end{figure}


\subsection{Mechanism Ablation}
\label{sec:ablation-studies}

The canonical EPD instantiation combines temporal accumulation, state-dependent release, and post-release reset. We next examine whether all three components are necessary for maintaining the empirical effectiveness of
the canonical dynamics. To this end, matched-control experiments using ResNet-18 with AdamW as a representative configuration are conducted in this section. Each matched control modifies one component of the canonical instantiation while keeping the remaining training and EPD settings unchanged. An additional Random control removes both temporal accumulation and state-dependent release and is included as a reference. Table~\ref{tab:mechanism-ablation} reports the final test performance of EPD
and the controls. Detailed per-seed results are
provided in Table~\ref{tab:mechanism-ablation-per-seed} of
Appendix~\ref{app:experimental-details}.

\begin{table}[t]
\centering
\caption{
Mechanism ablation results on CIFAR-100 with ResNet-18 and AdamW.
Results are reported as mean $\pm$ standard deviation over 10 paired random
seeds. $\Delta$ denotes the paired top-1 accuracy difference between each
control setting and canonical EPD (control $-$ EPD).
The three component-wise controls remove temporal accumulation
($\beta=0$), replace state-dependent release with random release, or remove
post-release reset, respectively. The joint control removes temporal
accumulation and replaces state-dependent release with random release while
retaining the same target release rate.
}
\label{tab:mechanism-ablation}
\vspace{0.6em}
\setlength{\tabcolsep}{7pt}
\renewcommand{\arraystretch}{1.10}
\begin{tabular}{lcc}
\hline
\multicolumn{1}{c}{\bf Control setting} &
\multicolumn{1}{c}{\bf Top-1 Accuracy (\%)} &
\multicolumn{1}{c}{\bf Paired $\boldsymbol{\Delta}$ (\%)} \\
\hline
Canonical EPD
& $73.65 \pm 0.19$
& -- \\

No temporal accumulation
& $66.89 \pm 0.40$
& $-6.76 \pm 0.34$ \\

Random release
& $71.63 \pm 0.23$
& $-2.02 \pm 0.30$ \\

No post-release reset
& $63.31 \pm 0.84$
& $-10.34 \pm 0.86$ \\

Random release without accumulation
& $52.54 \pm 0.36$
& $-21.11 \pm 0.37$ \\
\hline
\end{tabular}
\end{table}

\begin{figure}[t]
    \centering
    \includegraphics[width=\linewidth]{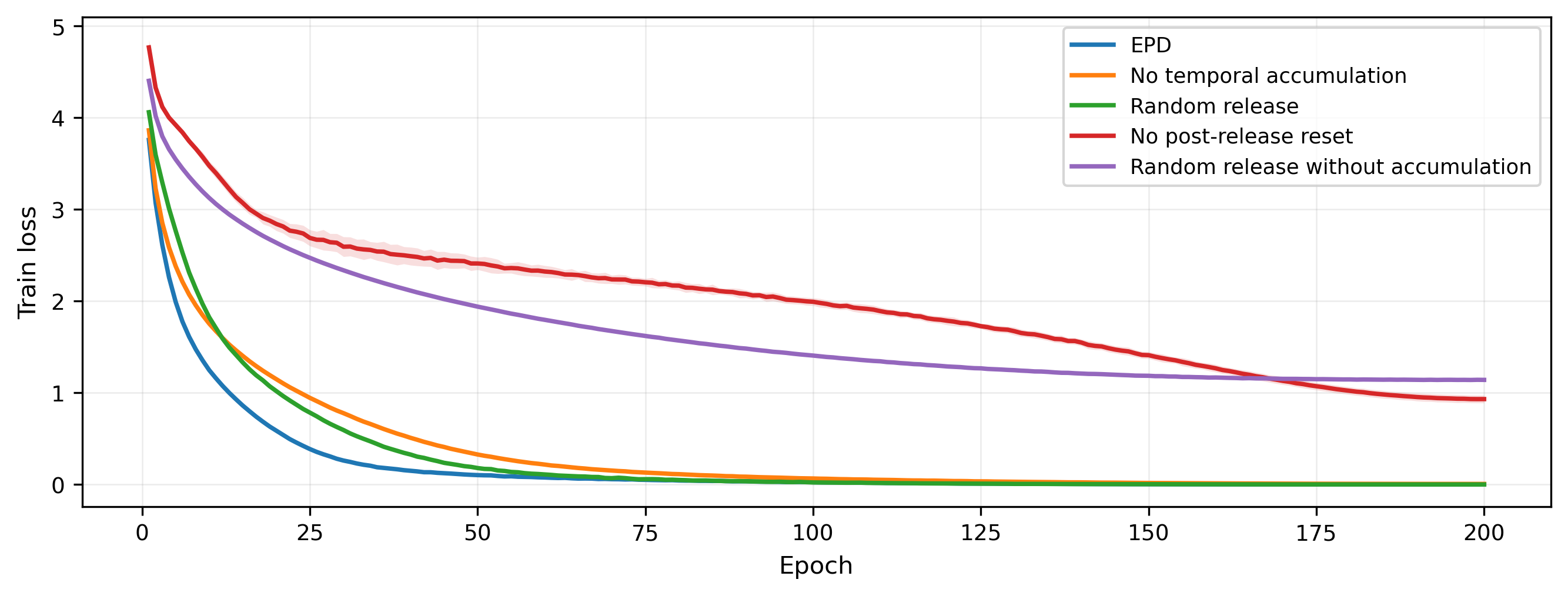}
    \caption{
    Training-loss trajectories for canonical EPD and the matched-control
    settings on CIFAR-100 with ResNet-18 and AdamW.
    Curves show the mean over 10 paired random seeds, and shaded regions
    indicate one standard deviation.
    }
    \label{fig:ablation-loss}
\end{figure}

\textbf{Temporal accumulation.}
The Top-k control removes temporal accumulation by setting $\beta=0$, while
retaining the same state-dependent Top-$p$ release rule and post-release reset. Removing accumulation reduces the mean top-1 accuracy from $73.65\%$ to $66.89\%$, corresponding to a paired difference of $-6.76 \pm 0.34$ percentage points. The consistent degradation across paired runs indicates that retaining candidate updates across optimization steps is
an important component of the canonical EPD dynamics in this setting. A broader sweep over nonzero values of $\beta$ further shows that the performance improves progressively as temporal retention is strengthened and becomes relatively stable in the high-$\beta$ regime, indicating that the benefit of accumulation is not tied to a single decay value. Detailed results are provided by ~\ref{tab:beta-sensitivity} and ~\ref{fig:beta-sensitivity}.

\textbf{State-dependent release.}
Random-Acc retains temporal accumulation and post-release reset, but replaces state-dependent Top-$p$ release rule with random release under the same target rate. This modification decreases the mean accuracy from $73.65\%$ to $71.63\%$, with a paired difference of $-2.02 \pm 0.30$ percentage points. The result indicates that retaining temporal accumulation alone is not sufficient to recover the performance of canonical EPD. State-dependent selection of which accumulated coordinates are released also contributes to its empirical effectiveness.

A joint control further removes temporal accumulation while replacing state-dependent release with random release, with the target release rate and post-release reset kept unchanged. The mean accuracy decreases to $52.54\%$, corresponding to a paired difference of
$-21.11 \pm 0.37$ percentage points relative to EPD. This result shows that maintaining the same release budget alone is insufficient to preserve the performance of the canonical dynamics, and further highlights the importance of how candidate updates are accumulated and selected for realization.

\textbf{Post-release reset.}
Post-release reset is removed while temporal accumulation and
state-dependent release are kept unchanged. Without reset, the mean top-1 accuracy decreases from $73.65\%$ to $63.31\%$, corresponding to a paired difference of $-10.34 \pm 0.86$ percentage points. The consistent degradation across paired runs indicates that resetting
the released state is an important component of the canonical EPD dynamics. Without this reset, released information remains in the accumulation state and continues to influence subsequent realization, substantially altering the intended accumulation--release process.

Figure~\ref{fig:ablation-loss} further compares the training trajectories under the matched-control settings. The controls exhibit qualitatively different effects on optimization. Removing temporal accumulation or replacing
state-dependent release with random release still allows the training loss to approach zero, despite their lower final test accuracy. In contrast, removing post-release reset or jointly removing temporal accumulation and state-dependent release results in substantially higher training loss over
much of the training process. These differences show that the matched controls alter not only final performance but also the optimization trajectories of the canonical dynamics.

Taken together, the matched controls show that the behavior of canonical EPD cannot be explained by partial parameter realization alone. Removing temporal accumulation, state-dependent release, or post-release reset individually
degrades final performance, while the corresponding training trajectories further show that these modifications affect optimization in different ways. Overall, the results indicate that all three components contribute to the
effectiveness of the canonical EPD dynamics.

\section{Conclusion and Discussion}
\label{sec:conclusion-and-discussion}

This work introduced Event-Driven Parameter Dynamics (EPD) as a framework that separates the generation of candidate parameter increments from their realization as parameter updates. The theoretical and empirical results together show that parameter realization can be treated as an explicit and controllable component of optimization dynamics. In the canonical instantiation, accumulation introduces temporal integration, release imposes budget-constrained spatial realization, and reset separates successive accumulation intervals, while conventional dense realization is recovered as a boundary case. Empirically, EPD maintains performance close to conventional training even when only a small fraction of parameter increments are realized at each step, including the $1\%$ release regime across the evaluated settings. The release-rate analysis and matched controls further show that the realization budget can be explicitly controlled and that accumulation, state-dependent release, and post-release reset each contribute to the resulting dynamics.

These results support a broader view of optimization design in which candidate-update generation and parameter realization provide complementary design dimensions. The timing, location, and persistence of parameter realization can be explicitly structured in addition to the generation of candidate updates. The canonical EPD studied here provides one concrete realization of this perspective through coordinate-wise accumulation, selective release, and reset. EPD explicitly models parameter realization as an independently structured and controllable process, thereby expanding the design space of optimization dynamics.

\textbf{Limitations.}
The present study covers CIFAR-100 with ResNet-18 and DenseNet-121 under AdamW and SGD-M, leaving broader architectures, datasets, and learning tasks to be evaluated. The theoretical analysis focuses on mechanism-level properties of the canonical EPD dynamics, while convergence, stability, and their relation to the overall optimization process remain open for further study. The current experiments primarily establish the functional roles of accumulation, state-dependent release, and reset, while the temporal and spatial organization of the resulting event patterns remains to be systematically characterized.

\textbf{Implications and future work.}
Future work should evaluate EPD across broader architectures and tasks, characterize the temporal and spatial structure of event-driven parameter realization, and develop optimization-level analyses of convergence, stability, and interactions with stochastic update dynamics. Beyond the canonical coordinate-wise instantiation, the EPD framework allows the parameter-unit granularity, accumulation state, event criterion, release rule, and reset mechanism to be adapted to different optimization objectives and system constraints. Exploring these extensions may reveal settings in which explicitly designed realization dynamics provide practical benefits beyond the current proof-of-concept study.

\subsection*{AI use statement}

In this work, we used generative AI tools to assist with research ideation and execution, including discussions and feedback on experimental design and analysis. Additionally, we used generative AI tools for literature retrieval and discovery, as well as language editing and polishing of the manuscript. And Generative AI tools were used to assist in formulating and checking mechanism-level mathematical claims and in drafting their derivations. All proofs and claims were reviewed by the authors. We did not use generative AI tools to generate synthetic datasets and other required disclosure tasks not relevant to this work are not applicable.

We have reviewed all AI-assisted work. Literature identified with AI assistance was independently checked against the original sources, and methodological suggestions, experimental designs, analyses, and interpretations were independently evaluated by the authors. All experimental results and scientific claims reported in the paper were verified by the authors. We take responsibility for the final content of this work, including text, claims, or artifacts produced with the aid of generative AI.

\subsection*{Reproducibility statement}

We provide the complete specification of the EPD update rule and its canonical coordinate-wise instantiation in Section~2. Experimental settings, including model architectures, optimizers, hyperparameters, release budgets, and random seeds, are described in Section~3 and the Appendix. Additional training curves, matched-control configurations, and implementation details are included in the supplementary material. If applicable, complete proofs of the theoretical properties are provided in the Appendix. Anonymous source code and configuration files are included in the supplementary materials to facilitate reproduction of the reported results.


\bibliography{iclr2027_conference}
\bibliographystyle{iclr2027_conference}

\clearpage
\appendix
\renewcommand{\thetable}{\thesection.\arabic{table}}
\renewcommand{\thefigure}{\thesection.\arabic{figure}}

\section{Algorithmic Description of the Canonical EPD Instantiation}
\label{app:algorithm}
\setcounter{algorithm}{0}

Algorithm~\ref{alg:epd-instantiation} summarizes the implementation of the canonical 
coordinate-wise EPD instantiation used in the experiments. The base optimizer computes dense 
candidate updates, while EPD accumulates these updates over time and releases only selected 
coordinates to the parameters according to the target release rate \(p\). Unreleased components 
remain in the accumulator for future steps.

\begin{algorithm}
\caption{Canonical EPD training instance}
\label{alg:epd-instantiation}
\begin{algorithmic}[1]
\Require initial parameters $\theta_0\in\mathbb{R}^d$, base optimizer state $h_0$, EPD states $\{s_0^{(k)}=0\}_{k=1}^d$, release rate $p$, accumulation decay $\beta$
\Ensure trained parameters $\theta_T$

\For{$t=0,1,\ldots,T-1$}
    \State Compute the mini-batch gradient
    \[
        g_t=\nabla_\theta \ell_t(\theta_t).
    \]

    \State Compute the optimizer-produced increment and update the optimizer state
    \[
        (u_t,h_{t+1})
        =
        \mathcal{B}(g_t,h_t,\theta_t).
    \]

    \For{$k=1,\ldots,d$}
        \State Accumulate the coordinate state
        \[
            \bar{s}_{t+1}^{(k)}
            =
            \beta s_t^{(k)}
            +
            u_t^{(k)}.
        \]
    \EndFor

    \For{each learnable parameter tensor $j$ with coordinate-index set $\mathcal{T}_j$}
    \State Set the tensor size and release budget
    \[
        d_j = |\mathcal{T}_j|,
        \qquad
        q_j =
        \max\left(
            1,
            \left\lfloor p d_j \right\rfloor
        \right).
    \]

    \State Select the $q_j$ coordinates with the largest accumulated-state magnitudes
    \[
        \mathcal{S}_{t+1}^{(j)}
        =
        \operatorname{Top}_{q_j}
        \left(
            \left\{
                \left|
                    \bar{s}_{t+1}^{(k)}
                \right|
            \right\}_{k\in\mathcal{T}_j}
        \right).
    \]

        \For{each coordinate $k\in\mathcal{T}_j$}
            \State Compute the release mask
            \[
                m_{t+1}^{(k)}
                =
                \mathds{1}
                \left(
                    k\in\mathcal{S}_{t+1}^{(j)}
                \right).
            \]

            \State Compute the realized increment
            \[
                r_{t+1}^{(k)}
                =
                m_{t+1}^{(k)}
                \bar{s}_{t+1}^{(k)}.
            \]

            \State Apply hard reset
            \[
                s_{t+1}^{(k)}
                =
                \left(
                    1-m_{t+1}^{(k)}
                \right)
                \bar{s}_{t+1}^{(k)}.
            \]
        \EndFor
    \EndFor

    \State Assemble the coordinate-wise realized increments into $r_{t+1}$.
    \State Update the parameters
    \[
        \theta_{t+1}
        =
        \theta_t
        +
        r_{t+1}.
    \]
\EndFor
\end{algorithmic}
\end{algorithm}

\clearpage
\section{Derivations for the EPD Properties}
\label{app:epd-properties}
\numberwithin{equation}{section}
\setcounter{equation}{0}

\paragraph{Proof of Proposition 1.}
From the general accumulation and reset rules,
\begin{equation}
\bar{s}_{t+1}^{(k)}
=
\mathcal{F}_{\mathrm{acc}}
\left(
\mathcal{R}
\left(
\bar{s}_{t}^{(k)},m_t^{(k)}
\right),
u_t^{(k)}
\right).
\label{eq:app-prop1-general}
\end{equation}
Thus, whenever the composition of $\mathcal{F}_{\mathrm{acc}}$ and
$\mathcal{R}$ retains a nontrivial dependence on the preceding state,
earlier candidate increments can propagate into subsequent accumulated
states.

For canonical EPD with hard reset,
\begin{equation}
\bar{s}_{t+1}^{(k)}
=
\beta
\left(
1-m_t^{(k)}
\right)
\bar{s}_{t}^{(k)}
+
u_t^{(k)}.
\label{eq:app-prop1-recurrence}
\end{equation}
Let $\rho$ denote the first optimization step after the preceding reset,
so that $s_{\rho}^{(k)}=0$. If no release occurs before the state
$\bar{s}_{t+1}^{(k)}$ is formed, recursive substitution gives
\begin{equation}
\bar{s}_{t+1}^{(k)}
=
\sum_{\ell=\rho}^{t}
\beta^{\,t-\ell}
u_{\ell}^{(k)}.
\label{eq:app-prop1-expansion}
\end{equation}

When $\beta=0$, only the current candidate increment remains. For
$0<\beta<1$, preceding candidate increments are retained with
exponentially decaying weights. When $\beta=1$, all candidate increments
within the current inter-release interval are retained without decay.
This establishes the temporal integration property.
\hfill$\square$

\paragraph{Proof of Corollary 1.}
Consider an inter-release interval containing $L$ candidate increments,
with
\begin{equation}
u_{\ell}^{(k)}
=
\mu^{(k)}
+
\varepsilon_{\ell}^{(k)},
\label{eq:app-noise-model}
\end{equation}
where $\mu^{(k)}$ is constant within the interval,
$\mathbb{E}[\varepsilon_{\ell}^{(k)}]=0$, and the perturbations are
mutually uncorrelated with
$\operatorname{Var}(\varepsilon_{\ell}^{(k)})=\sigma_k^2$.

From Proposition~1, the accumulated state can be written as
\begin{equation}
\bar{s}^{(k)}
=
\sum_{h=0}^{L-1}
\beta^h
\left(
\mu^{(k)}+\varepsilon_h^{(k)}
\right).
\label{eq:app-noise-accumulation}
\end{equation}
Define the normalized accumulated state as
\begin{equation}
\widetilde{s}^{(k)}
=
\frac{
\bar{s}^{(k)}
}{
\sum_{h=0}^{L-1}\beta^h
}.
\label{eq:app-normalized-state}
\end{equation}
Its expectation is
\begin{equation}
\mathbb{E}
\left[
\widetilde{s}^{(k)}
\right]
=
\mu^{(k)}.
\label{eq:app-noise-expectation}
\end{equation}

Since the perturbations are mutually uncorrelated,
\begin{equation}
\operatorname{Var}
\left[
\widetilde{s}^{(k)}
\right]
=
\sigma_k^2
\frac{
\sum_{h=0}^{L-1}\beta^{2h}
}{
\left(
\sum_{h=0}^{L-1}\beta^h
\right)^2
}.
\label{eq:app-noise-variance}
\end{equation}
For $\beta\in[0,1]$,
\begin{equation}
\sum_{h=0}^{L-1}\beta^{2h}
\leq
\left(
\sum_{h=0}^{L-1}\beta^h
\right)^2,
\end{equation}
and therefore
\begin{equation}
\operatorname{Var}
\left[
\widetilde{s}^{(k)}
\right]
\leq
\sigma_k^2.
\label{eq:app-noise-bound}
\end{equation}
Equality holds for $\beta=0$, while the inequality is strict for
$\beta>0$ when more than one nonzero contribution is present. Hence,
under the stated local-consistency assumptions, normalized temporal
accumulation preserves the persistent component while attenuating
uncorrelated transient perturbations.
\hfill$\square$

\paragraph{Proof of Proposition 2.}
For general EPD,
\begin{equation}
r_{t+1}
=
m_{t+1}
\odot
\bar{s}_{t+1}.
\label{eq:app-prop2-realization}
\end{equation}
Any coordinate with $m_{t+1}^{(k)}=0$ therefore satisfies
$r_{t+1}^{(k)}=0$, which gives
\begin{equation}
\operatorname{supp}(r_{t+1})
\subseteq
\operatorname{supp}(m_{t+1}),
\qquad
\|r_{t+1}\|_0
\leq
\|m_{t+1}\|_0.
\label{eq:app-prop2-support}
\end{equation}

For tensor $\mathcal{T}_j$, canonical EPD selects $q_j$ coordinates
with the largest accumulated-state magnitudes. Hence,
\begin{equation}
\mathcal{S}_{t+1}^{(j)}
\in
\arg\max_{\substack{
\mathcal{S}\subseteq\mathcal{T}_j\\
|\mathcal{S}|=q_j
}}
\sum_{k\in\mathcal{S}}
\left|
\bar{s}_{t+1}^{(k)}
\right|^2.
\label{eq:app-prop2-topq}
\end{equation}
For any support $\mathcal{S}\subseteq\mathcal{T}_j$ with
$|\mathcal{S}|\leq q_j$, the minimum approximation error over vectors
supported on $\mathcal{S}$ is obtained by retaining the corresponding
entries of $\bar{s}_{t+1}^{(j)}$. The resulting error is
\begin{equation}
\sum_{k\in\mathcal{T}_j\setminus\mathcal{S}}
\left|
\bar{s}_{t+1}^{(k)}
\right|^2.
\label{eq:app-prop2-error}
\end{equation}
Minimizing this error is equivalent to maximizing the retained squared
magnitude. The Top-$q_j$ support therefore satisfies
\begin{equation}
r_{t+1}^{(j)}
\in
\arg\min_{\substack{
z\\
\|z\|_0\leq q_j
}}
\left\|
\bar{s}_{t+1}^{(j)}-z
\right\|_2^2.
\label{eq:app-prop2-best-sparse}
\end{equation}
Thus, the canonical release rule gives a best $q_j$-sparse
approximation of the accumulated state within each tensor.
\hfill$\square$

\paragraph{Proof of Corollary 2.}
For canonical EPD, setting $p=1$ gives
\begin{equation}
q_j=d_j
\end{equation}
for every tensor $\mathcal{T}_j$. Hence, every coordinate is selected
for release at every step and
\begin{equation}
m_{t+1}^{(k)}=1
\end{equation}
for all $k$.

Under hard reset,
\begin{equation}
s_{t+1}^{(k)}
=
\left(
1-m_{t+1}^{(k)}
\right)
\bar{s}_{t+1}^{(k)}
=
0.
\label{eq:app-cor2-reset}
\end{equation}
Together with the initial condition $s_0=0$, this gives $s_t=0$ at
every step. The canonical accumulation rule therefore reduces to
\begin{equation}
\bar{s}_{t+1}
=
u_t.
\label{eq:app-cor2-accumulation}
\end{equation}
Since every coordinate is released,
\begin{equation}
r_{t+1}
=
m_{t+1}
\odot
\bar{s}_{t+1}
=
u_t.
\label{eq:app-cor2-realization}
\end{equation}
The parameter update consequently becomes
\begin{equation}
\theta_{t+1}
=
\theta_t
+
u_t,
\label{eq:app-cor2-conventional}
\end{equation}
which is exactly the conventional dense realization of the base
optimizer. This establishes conventional dense realization as a
boundary case of canonical EPD.
\hfill$\square$

\paragraph{Proof of Proposition 3.}
For general EPD, the state entering the next accumulation step is
determined by the reset rule,
\begin{equation}
s_t^{(k)}
=
\mathcal{R}
\left(
\bar{s}_{t}^{(k)},m_t^{(k)}
\right).
\label{eq:app-prop3-general-reset}
\end{equation}
Canonical hard reset gives
\begin{equation}
s_t^{(k)}
=
\left(
1-m_t^{(k)}
\right)
\bar{s}_{t}^{(k)}.
\label{eq:app-prop3-reset}
\end{equation}
Substituting this relation into the canonical accumulation rule yields
\begin{equation}
\bar{s}_{t+1}^{(k)}
=
\beta
\left(
1-m_t^{(k)}
\right)
\bar{s}_{t}^{(k)}
+
u_t^{(k)}.
\label{eq:app-prop3-recurrence}
\end{equation}

Repeated substitution for $n$ steps gives
\begin{equation}
\begin{aligned}
\bar{s}_{t+n}^{(k)}
={}&
\beta^n
\left[
\prod_{r=0}^{n-1}
\left(
1-m_{t+r}^{(k)}
\right)
\right]
\bar{s}_{t}^{(k)}
\\
&+
\sum_{j=0}^{n-1}
\beta^{\,n-1-j}
\left[
\prod_{r=j+1}^{n-1}
\left(
1-m_{t+r}^{(k)}
\right)
\right]
u_{t+j}^{(k)} .
\end{aligned}
\label{eq:app-prop3-expansion}
\end{equation}
An empty product is understood as one.

Since $m_{t+r}^{(k)}\in\{0,1\}$, a release at any subsequent step makes
the corresponding product containing
$1-m_{t+r}^{(k)}$ equal to zero. Contributions accumulated before that
release therefore do not propagate beyond it.

In particular, when $m_t^{(k)}=1$,
Eq.~\ref{eq:app-prop3-recurrence} reduces to
\begin{equation}
\bar{s}_{t+1}^{(k)}
=
u_t^{(k)}.
\label{eq:app-prop3-release}
\end{equation}
When $m_t^{(k)}=0$,
\begin{equation}
\bar{s}_{t+1}^{(k)}
=
\beta
\bar{s}_{t}^{(k)}
+
u_t^{(k)},
\label{eq:app-prop3-no-release}
\end{equation}
so the preceding accumulated state continues to propagate. Hard reset
therefore separates the temporal history of each coordinate into
successive inter-release accumulation intervals.
\hfill$\square$

\clearpage
\section{Experimental Details}
\label{app:experimental-setting}
\setcounter{table}{0}
\setcounter{figure}{0}

This appendix provides the training and implementation settings used in the empirical evaluation, together with the EPD-specific settings of the canonical instantiation. Unless otherwise specified, conventional training and the corresponding EPD runs use the same training protocol and experimental environment. Settings that are modified in specific analyses or ablation studies are indicated where relevant.

\subsection{Training and Implementation Settings}
\label{app:training-hyperparameters}

Unless otherwise specified, conventional training and the corresponding EPD runs use the same training protocol. 
The empirical evaluation is conducted on CIFAR-100 using ResNet-18 and DenseNet-121 with two base optimizers, AdamW and SGD with momentum (SGD-M). 
The common training settings and optimizer-specific hyperparameters are summarized in Table~\ref{tab:training-optimization-hyperparameters}.

\begin{table}[ht]
\caption{Training and optimization settings used in the empirical evaluation.}
\label{tab:training-optimization-hyperparameters}
\begin{center}
\setlength{\tabcolsep}{7pt}
\renewcommand{\arraystretch}{1.15}

\begin{tabular}{lcc}
\hline
\multicolumn{1}{c}{\bf Item} &
\multicolumn{1}{c}{\bf AdamW} &
\multicolumn{1}{c}{\bf SGD-M} \\
\hline

Dataset
& \multicolumn{2}{c}{CIFAR-100} \\[2pt]

Models
& \multicolumn{2}{c}{ResNet-18, DenseNet-121} \\[2pt]

Training epochs
& \multicolumn{2}{c}{200} \\[2pt]

Batch size
& \multicolumn{2}{c}{128} \\[2pt]

Random seeds
& \multicolumn{2}{c}{10} \\[2pt]

Data augmentation
& \multicolumn{2}{c}{Random crop with padding, random horizontal flip, normalization} \\[2pt]

Learning-rate schedule
& \multicolumn{2}{c}{Cosine annealing} \\

\hline

Initial learning rate
& 0.001
& 0.1 \\[2pt]

Optimizer coefficients
& $\beta_1=0.9,\ \beta_2=0.999$
& momentum $=0.9$ \\[2pt]

Weight decay
& 0.05
& 0.0005 \\

\hline
\end{tabular}
\end{center}
\end{table}

All experiments are implemented in PyTorch 2.5.1 with CUDA 12.4 and conducted on workstations equipped with NVIDIA RTX A6000 GPUs. 
For each controlled comparison, conventional training and the corresponding EPD run use the same software and hardware environment.

\subsection{Canonical EPD Settings}
\label{app:epd-hyperparameters}

All EPD experiments are based on the canonical coordinate-wise instantiation described in Section~\ref{sec:canonical-instantiation}. 
Unless otherwise specified, the EPD-specific settings listed in Table~\ref{tab:epd-hyperparameters} are used throughout the experiments.

\begin{table}[ht]
\caption{Settings of the canonical EPD instantiation.}
\label{tab:epd-hyperparameters}
\begin{center}
\setlength{\tabcolsep}{7pt}
\renewcommand{\arraystretch}{1.15}
\begin{tabular}{lll}
\hline
\multicolumn{1}{c}{\bf Item} &
\multicolumn{1}{c}{\bf Default value} &
\multicolumn{1}{c}{\bf Usage} \\
\hline

Accumulation decay $\beta$
& 0.95
& Fixed unless otherwise specified \\

Target release rate $p$
& 1\%
& Varied in release-control experiments \\

Reset rule
& Hard reset
& Modified only in mechanism ablations \\

\hline
\end{tabular}
\end{center}
\end{table}

\clearpage
\section{Detailed experimental results}
\label{app:experimental-details}
\setcounter{table}{0}
\setcounter{figure}{0}

This section provides additional results for the main experiments in 
Section~\ref{sec:main-results}. Table~\ref{tab:per-seed-results} reports the per-seed top-1 
accuracy corresponding to the main results in Table~\ref{tab:main-results}. Each row corresponds to one random seed, and each pair of columns compares training without EPD and training with the canonical EPD instantiation under the same model--optimizer setting.

\begin{table}[ht]
\caption{Per-seed results on CIFAR-100. Top-1 test accuracy is reported for each random seed. 
Each model--optimizer setting compares conventional training without EPD and the canonical EPD instantiation. Here, EPD denotes the canonical coordinate-wise instantiation. R18 denotes ResNet-18, DN121 denotes DenseNet-121, SGD-M denotes SGD with momentum, and w/o denotes without EPD.}
\label{tab:per-seed-results}
\vspace{0.6em}
\begin{center}
\setlength{\tabcolsep}{8.0pt}
\renewcommand{\arraystretch}{1.15}
\small
\begin{tabular}{c cc cc cc cc}
\multicolumn{1}{c}{\multirow{3}{*}{\bf Seed}}
& \multicolumn{8}{c}{\bf Top-1 Test Acc. (\%)} \\
\cline{2-9}
& \multicolumn{2}{c}{\bf R18 + AdamW} 
& \multicolumn{2}{c}{\bf DN121 + AdamW} 
& \multicolumn{2}{c}{\bf R18 + SGD-M} 
& \multicolumn{2}{c}{\bf DN121 + SGD-M} \\
& \multicolumn{1}{c}{\bf w/o} & \multicolumn{1}{c}{\bf EPD} 
& \multicolumn{1}{c}{\bf w/o} & \multicolumn{1}{c}{\bf EPD} 
& \multicolumn{1}{c}{\bf w/o} & \multicolumn{1}{c}{\bf EPD} 
& \multicolumn{1}{c}{\bf w/o} & \multicolumn{1}{c}{\bf EPD} \\
\hline

0 & 73.83 & 73.95 & 76.74 & 76.39 & 78.67 & 77.27 & 80.64 & 80.08 \\
1 & 73.45 & 73.73 & 76.67 & 76.18 & 78.65 & 76.99 & 80.53 & 79.77 \\
2 & 74.00 & 73.93 & 76.86 & 76.20 & 78.39 & 76.87 & 79.89 & 79.85 \\
3 & 74.34 & 73.43 & 76.97 & 76.05 & 79.03 & 77.16 & 80.38 & 80.25 \\
4 & 74.11 & 73.63 & 76.79 & 75.57 & 78.80 & 77.38 & 80.76 & 79.70 \\
5 & 74.20 & 73.33 & 76.91 & 76.14 & 78.94 & 77.52 & 80.78 & 79.38 \\
6 & 73.84 & 73.55 & 77.06 & 76.05 & 79.00 & 77.40 & 80.62 & 79.85 \\
7 & 74.34 & 73.68 & 76.61 & 75.93 & 79.30 & 77.20 & 80.69 & 79.87 \\
8 & 73.77 & 73.60 & 77.35 & 75.71 & 78.42 & 76.86 & 80.87 & 79.56 \\
9 & 74.07 & 73.68 & 77.10 & 75.63 & 78.43 & 77.50 & 80.65 & 80.24 \\

\hline

Mean 
& 74.00 & 73.65
& 76.91 & 75.98
& 78.76 & 77.22
& 80.58 & 79.85 \\

Std. 
& 0.28 & 0.19
& 0.22 & 0.27
& 0.31 & 0.24
& 0.28 & 0.28 \\

\hline
\end{tabular}
\end{center}
\end{table}

Tables~\ref{tab:release-rate-per-seed} and
\ref{tab:mechanism-ablation-per-seed} provide the detailed per-seed results for the release-rate analysis and the matched-control ablations, respectively. Both analyses are conducted on CIFAR-100 with ResNet-18 and AdamW as the representative configuration. The former corresponds to the release-rate results summarized in Figure~\ref{fig:main-curves}(e), while the latter provides the per-seed results underlying Table~\ref{tab:mechanism-ablation}.

\begin{table}[ht]
\caption{Per-seed results under different release rates on CIFAR-100 with ResNet-18 and AdamW. 
Top-1 test accuracy is reported for each random seed. 
The release rate \(p\) denotes the fraction of candidate updates released at each training step.}
\label{tab:release-rate-per-seed}
\vspace{0.6em}
\begin{center}
\setlength{\tabcolsep}{4.5pt}
\renewcommand{\arraystretch}{1.15}
\footnotesize
\begin{tabular}{c c c c c c c c c c}
\multicolumn{1}{c}{\multirow{2}{*}{\bf Seed}}
& \multicolumn{9}{c}{\bf Top-1 Test Acc. (\%)} \\
\cline{2-10}
& \multicolumn{1}{c}{\(\boldsymbol{0.5\%}\)}
& \multicolumn{1}{c}{\(\boldsymbol{1\%}\)}
& \multicolumn{1}{c}{\(\boldsymbol{5\%}\)}
& \multicolumn{1}{c}{\(\boldsymbol{10\%}\)}
& \multicolumn{1}{c}{\(\boldsymbol{20\%}\)}
& \multicolumn{1}{c}{\(\boldsymbol{40\%}\)} 
& \multicolumn{1}{c}{\(\boldsymbol{60\%}\)}
& \multicolumn{1}{c}{\(\boldsymbol{80\%}\)} 
& \multicolumn{1}{c}{\(\boldsymbol{100\%}\)} 
\\
\hline
0 & 73.06 & 73.95 & 73.40 & 73.86 & 74.06 & 73.83 & 73.95 & 74.10 & 73.84 \\
1 & 73.51 & 73.73 & 73.81 & 73.92 & 73.59 & 73.76 & 73.86 & 73.46 & 73.72 \\
2 & 73.88 & 73.93 & 73.28 & 73.99 & 74.00 & 73.76 & 74.23 & 74.27 & 73.95 \\
3 & 72.95 & 73.43 & 73.10 & 73.61 & 73.53 & 73.42 & 73.85 & 73.71 & 73.56 \\
4 & 73.12 & 73.63 & 73.79 & 74.37 & 74.04 & 74.46 & 73.79 & 74.14 & 74.04 \\
\hline
Mean
& 73.30 & 73.73 & 73.48 & 73.95 & 73.84 & 73.85 & 73.94 & 73.94 & 73.82 \\
Std.
& 0.39 & 0.22 & 0.31 & 0.28 & 0.26 & 0.38 & 0.17 & 0.34 & 0.19 \\
\hline
\end{tabular}
\end{center}
\end{table}

\begin{table}[ht]
\caption{
Per-seed results for the mechanism ablation experiments on CIFAR-100 with ResNet-18 and AdamW. Top-1 test accuracy is reported for each random seed. Here, EPD denotes the canonical EPD instantiation. Three matched controls modify one component at a time: Top-k removes temporal accumulation by setting
$\beta=0$, Random-Acc replaces state-dependent release with random release while retaining accumulation, and No-Reset removes the post-release reset. An additional Random control removes both temporal accumulation and state-dependent release as a reference.}
\label{tab:mechanism-ablation-per-seed}
\vspace{0.6em}
\begin{center}
\setlength{\tabcolsep}{7pt}
\renewcommand{\arraystretch}{1.15}
\footnotesize
\begin{tabular}{c c c c c c}
\multicolumn{1}{c}{\multirow{2}{*}{\bf Seed}}
& \multicolumn{5}{c}{\bf Top-1 Test Acc. (\%)} \\
\cline{2-6}
& \multicolumn{1}{c}{\bf EPD}
& \multicolumn{1}{c}{\bf Top-k ($\boldsymbol{\beta=0}$)}
& \multicolumn{1}{c}{\bf Random-Acc}
& \multicolumn{1}{c}{\bf No-Reset}
& \multicolumn{1}{c}{\bf Random} \\
\hline
0 & 73.95 & 67.34 & 71.59 & 63.09 & 52.61 \\
1 & 73.73 & 67.69 & 71.63 & 61.85 & 52.90 \\
2 & 73.93 & 67.24 & 71.90 & 64.86 & 52.72 \\
3 & 73.43 & 66.86 & 71.91 & 63.75 & 52.95 \\
4 & 73.63 & 66.48 & 71.74 & 62.54 & 52.15 \\
5 & 73.33 & 66.72 & 71.45 & 63.96 & 52.10 \\
6 & 73.55 & 66.74 & 71.85 & 62.75 & 53.04 \\
7 & 73.68 & 66.65 & 71.17 & 63.46 & 52.48 \\
8 & 73.60 & 66.62 & 71.55 & 63.11 & 52.16 \\
9 & 73.68 & 66.54 & 71.48 & 63.73 & 52.27 \\
\hline
Mean & 73.65 & 66.89 & 71.63 & 63.31 & 52.54 \\
Std. & 0.19 & 0.40 & 0.23 & 0.84 & 0.36 \\
\hline
\end{tabular}
\end{center}
\end{table}

\clearpage
\section{Sensitivity to the Accumulation Decay}
\label{app:beta-sensitivity}
\setcounter{table}{0}
\setcounter{figure}{0}

This appendix reports an additional analysis of temporal accumulation in the canonical EPD instantiation by examining sensitivity to the accumulation decay $\beta$. The analysis is conducted on CIFAR-100 with ResNet-18 and AdamW as the representative configuration.


The matched-control experiment with $\beta=0$ examines the effect of removing
temporal accumulation entirely. Here, we further vary $\beta$ over a broader
range to examine how the strength of temporal retention affects the canonical
EPD instantiation. All experiments use the same target release rate
$p=1\%$ and hard reset, while only $\beta$ is varied.

\begin{table}[ht]
\caption{Per-seed results for different accumulation decay values under
the main CIFAR-100, ResNet-18, and AdamW setting. All experiments use
the release rate \(p=1\%\) and hard reset.}
\label{tab:beta-sensitivity}
\begin{center}
\setlength{\tabcolsep}{4.5pt}
\renewcommand{\arraystretch}{1.15}
\footnotesize
\begin{tabular}{c c c c c c c c}
\multicolumn{1}{c}{\multirow{2}{*}{\bf Seed}}
& \multicolumn{7}{c}{\bf Top-1 Test Acc. (\%)} \\
\cline{2-8}
& \multicolumn{1}{c}{\(\boldsymbol{\beta=0}\)}
& \multicolumn{1}{c}{\(\boldsymbol{\beta=0.2}\)}
& \multicolumn{1}{c}{\(\boldsymbol{\beta=0.5}\)}
& \multicolumn{1}{c}{\(\boldsymbol{\beta=0.7}\)}
& \multicolumn{1}{c}{\(\boldsymbol{\beta=0.9}\)}
& \multicolumn{1}{c}{\(\boldsymbol{\beta=0.95}\)}
& \multicolumn{1}{c}{\(\boldsymbol{\beta=0.99}\)} \\
\hline
0 & 67.34 & 67.40 & 69.38 & 71.17 & 73.12 & 73.47 & 73.71 \\
1 & 67.69 & 67.82 & 69.37 & 71.34 & 73.22 & 73.62 & 73.69 \\
2 & 67.24 & 67.14 & 69.30 & 70.57 & 73.46 & 73.46 & 73.47 \\
3 & 66.86 & 67.40 & 69.30 & 71.42 & 73.26 & 73.87 & 73.30 \\
4 & 66.48 & 67.64 & 69.12 & 71.19 & 73.70 & 73.68 & 73.29 \\
5 & 66.72 & 67.66 & 69.63 & 70.81 & 73.43 & 73.90 & 73.47 \\
6 & 66.74 & 67.39 & 69.42 & 70.69 & 73.21 & 73.88 & 73.30 \\
7 & 66.65 & 67.52 & 68.72 & 71.18 & 73.24 & 73.35 & 73.50 \\
8 & 66.62 & 67.51 & 69.35 & 71.20 & 73.44 & 73.85 & 73.25 \\
9 & 66.54 & 67.21 & 69.35 & 70.91 & 73.15 & 73.88 & 73.95 \\
\hline
Mean & 66.89 & 67.47 & 69.29 & 71.05 & 73.32 & 73.70 & 73.49 \\
Std. & 0.40 & 0.21 & 0.24 & 0.29 & 0.18 & 0.21 & 0.23 \\
\hline
\end{tabular}
\end{center}
\end{table}

\begin{figure}[ht]
\begin{center}
\includegraphics[width=0.80\linewidth]{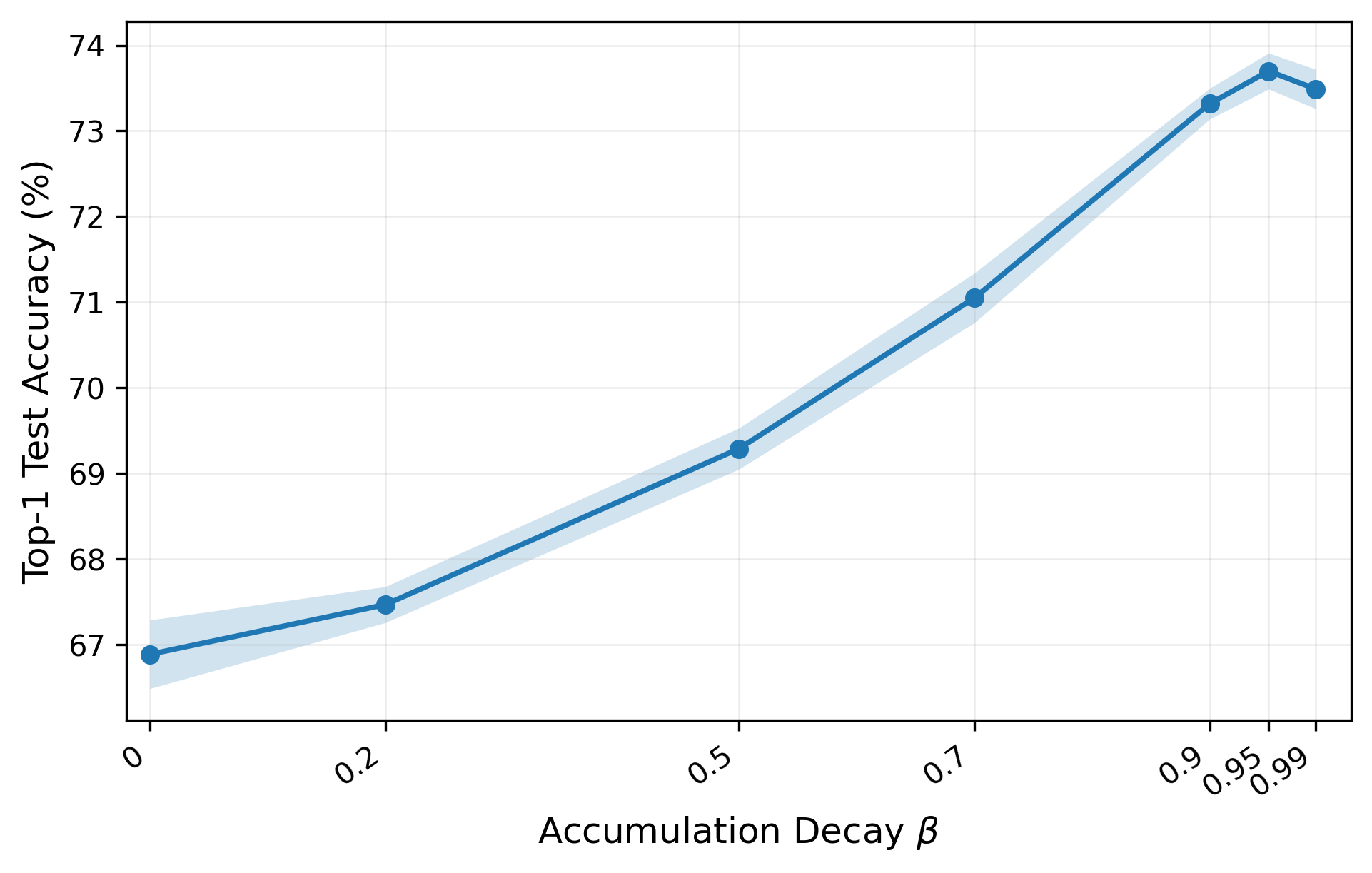}
\caption{
Top-1 test accuracy under different accumulation decay values $\beta$
on CIFAR-100 with ResNet-18 and AdamW.
The horizontal positions correspond to the actual numerical values of $\beta$.
The solid curve shows the mean over 10 random seeds, and the shaded region
indicates one standard deviation.
}
\label{fig:beta-sensitivity}
\end{center}
\end{figure}

Table~\ref{tab:beta-sensitivity} and Figure~\ref{fig:beta-sensitivity}
show a clear dependence on the strength of temporal retention. Performance
increases substantially as $\beta$ is raised from $0$ to the high-retention
regime, reaching $73.32\%$, $73.70\%$, and $73.49\%$ for
$\beta=0.9$, $0.95$, and $0.99$, respectively. In contrast, weaker retention
at $\beta=0.2$, $0.5$, and $0.7$ results in noticeably lower accuracy.
Within the tested high-$\beta$ range, however, the final performance remains
similar rather than continuing to improve monotonically. These results show that the performance of the canonical EPD instantiation
depends strongly on the degree of temporal retention when $\beta$ is small
or moderate. Once $\beta$ enters the high-retention regime, however, the
final accuracy becomes relatively stable across the tested values. Thus,
strong temporal accumulation appears important in this setting, while the
observed performance does not depend on tuning $\beta$ to a single precise
value.

\end{document}